\documentclass{isprs} 
\usepackage{subfigure}
\usepackage{setspace}
\usepackage{geometry} 
\usepackage{epstopdf}
\usepackage[labelsep=period]{caption}  
\usepackage[british]{babel} 
\usepackage[hang]{footmisc}

\usepackage[backend=biber,
style=numeric-comp,
natbib=true,
maxcitenames=2,
mincitenames=1,
sorting=none]{biblatex}
\usepackage{tabularx}
\newcolumntype{Y}{>{\centering\arraybackslash}X}
\usepackage{booktabs}
\usepackage{ragged2e} 
\usepackage[utf8]{inputenc} 
\usepackage[T1]{fontenc}    
\usepackage{hyperref}       
\usepackage{url}            
\usepackage{booktabs}       
\usepackage{amsfonts}       
\usepackage{nicefrac}       
\usepackage{microtype}      
\usepackage{tabularray}
\usepackage{algorithm}
\usepackage[noend]{algpseudocode}
\usepackage{dirtree}
\usepackage{graphicx}
\usepackage{caption}
\usepackage{amsmath}
\usepackage{cleveref}
\usepackage{comment}

\usepackage{tcolorbox}
\usepackage{multirow}
\usepackage{multicol}
\usepackage{setspace}
\usepackage{colortbl}
\usepackage{graphicx}

\begin{document}

\title{AGBD: A Global-scale Biomass Dataset}
\date{}


\author{
Ghjulia Sialelli\textsuperscript{1}, Torben Peters\textsuperscript{1}, Jan D. Wegner\textsuperscript{2}, Konrad Schindler\textsuperscript{1}}

\address{
	\textsuperscript{1 }Photogrammetry and Remote Sensing, ETH Zurich - (gsialelli, tpeters, schindler)@ethz.ch\\
	\textsuperscript{2 }EcoVision Lab, DM3L, University of Zurich - jandirk.wegner@uzh.ch\\
}



\abstract{

Accurate estimates of Above Ground Biomass (AGB) are essential in addressing two of humanity's biggest challenges: climate change and biodiversity loss. Existing datasets for AGB estimation from satellite imagery are limited. Either they focus on specific, local regions at high resolution, or they offer global coverage at low resolution. There is a need for a machine learning-ready, globally representative, high-resolution benchmark dataset. Our findings indicate significant variability in biomass estimates across different vegetation types, emphasizing the necessity for a dataset that accurately captures global diversity. To address these gaps, we introduce a comprehensive new dataset that is globally distributed, covers a range of vegetation types, and spans several years. This dataset combines AGB reference data from the GEDI mission with data from Sentinel-2 and PALSAR-2 imagery. Additionally, it includes pre-processed high-level features such as a dense canopy height map, an elevation map, and a land-cover classification map. We also produce a dense, high-resolution ($10\,$m) map of AGB predictions for the entire area covered by the dataset. Rigorously tested, our dataset is accompanied by several benchmark models and is publicly available. It can be easily accessed using a single line of code, offering a solid basis for efforts towards global AGB estimation. 

}


\maketitle


\section{Introduction}
 
\sloppy

Measuring Above Ground Biomass (AGB) is crucial to combat biodiversity loss and address the ongoing climate crisis. Accurate AGB estimates enable the quantification of carbon stocks, which play a pivotal role in carbon offsetting schemes. Furthermore, AGB is correlated with various biodiversity metrics, providing valuable insights into the structure of biodiversity hotspots \cite{VanceChalcraft2010, Sonkoly2019,Li2018}. Traditional methods for measuring AGB are based on field work, either via destructive sampling, where sample trees are cut and weighed; or with non-destructive techniques that use tree dimensions (like tree height and trunk diameter at breast height) in tailored regression equations. Increasingly, LiDAR campaigns are replacing manual measurements with Remote Sensing (RS), becoming the preferred method. However, despite their precision, the high costs and time requirements limit their global scalability. As a result, previous work has mainly focused on localized AGB estimation. Launched in 2019, the NASA GEDI mission uses an on-orbit laser altimeter aboard the International Space Station to scan the globe between $51.6^\circ$ north and south. Calibrated on coincident airborne LiDAR data and ground plot field inventories, the mission provides sparse AGB estimates across its observation domain. These estimates represent the largest existing reference AGB dataset, which can be combined with Machine Learning (ML) methods and RS data to advance AGB estimation across the globe. We provide a ML-ready dataset consisting of coincident GEDI AGB estimates and various RS data products. Although freely accessible, obtaining remotely sensed data globally is tedious, and the sheer scale of it is prohibitive for ML research. The corresponding global dataset would require approximately 70TB of storage, $\approx60$ times more than the complete ImageNet. For AGB estimation that data will need extensive preprocessing, and it would be too large to repeatedly train and test within reasonable time during method development.
To reduce the scale, but to the greatest possible extent maintain the representativeness of the dataset, we derive a subset of regions whose vegetation distribution emulates the global one. The complex interactions between AGB and various vegetation types \cite{Chen2023} motivate this choice. 

The dataset serves multiple purposes. (i) \textbf{Providing a globally representative high-res testbed for AGB estimation:} We show that all existing datasets are either too localized to generalize around the globe, or have only low resolution. Our collection covers all biomes and is globally representative at high resolution, enabling the training of AGB estimation models with improved performance and better generalization abilities.
(ii) \textbf{Improving Regional Performance:} Prior research has shown that combining GEDI data with local reference data yields better results than using only the local data alone \cite{Bullock2023}. Researchers can leverage our comprehensive dataset for initial training and then fine-tune their models with local data specific to their regions, enhancing the accuracy and performance of their analyses.

To highlight the accessibility of our dataset, we offer a fully preprocessed version that is compatible with all major ML frameworks, such as TensorFlow and PyTorch. It is hosted on \href{https://huggingface.co/datasets/prs-eth/AGBD}{HuggingFace} \cite{lhoest-etal-2021-datasets} and can be downloaded and used with just the following lines of code: 

\begin{figure}[ht]
\centering
\includegraphics[width=\linewidth]{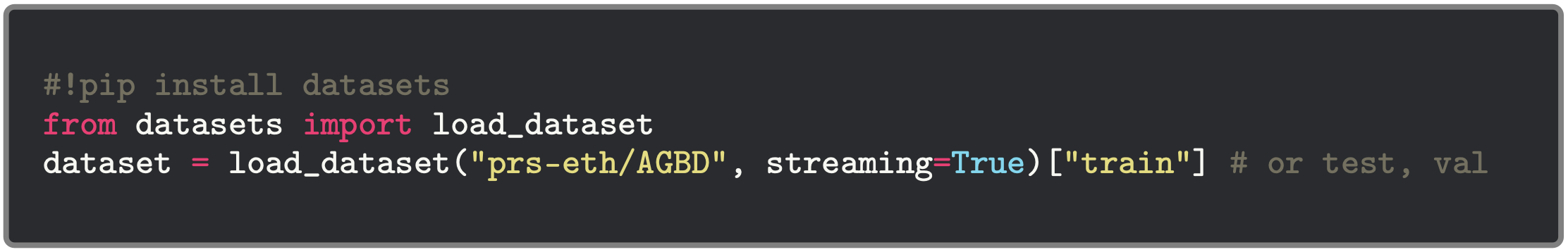}
\end{figure}

\newenvironment{myitemize}
{ \begin{itemize}
    \setlength{\itemsep}{0pt}
    \setlength{\parskip}{0pt}
    \setlength{\parsep}{0pt}     }
{ \end{itemize}                  } 

Our contributions are:
\vspace{-0.5em}
\begin{myitemize}
\item We offer a machine learning-ready, easily accessible, globally representative dataset comprising coincident AGB estimates together with various remote sensing data products.
\item We conduct a comprehensive analysis of our dataset and apply several standard models to it to validate its accuracy and reliability and to serve as baselines.
\item We generate and provide a dense, high-resolution ($10\,$m) map of AGB predictions across the dataset's coverage area with the best-performing baseline.
\item We release all benchmark models and pretrained weights.
\end{myitemize}

All code is hosted on \href{https://github.com/ghjuliasialelli/AGBD}{Github}, moreover the original raw data, the dense predictions, and the model weights are hosted on the \href{https://www.research-collection.ethz.ch/handle/20.500.11850/674193}{ETH Research Collection}.

\section{Background}
\label{sec:background}
In recent years, remote sensing (RS) and machine learning (ML) have advanced various mapping tasks, ranging from human population \cite{Metzger23} to canopy height \cite{Lang2023}. Biomass estimation also has seen significant research, with \cite{Zhang2019, Araza2022} providing a review of existing AGB datasets and maps, supplemented by Table \ref{tab:recent_work}. Currently, global-scale AGB datasets suffer from low spatial resolution (at best 100 meters), while high-resolution datasets (10-30 meters) are limited to restricted geographic areas. This dichotomy forces researchers to strike a compromise between generalization across space and spatial resolution. Notably, \cite{Araza2022} highlight the challenge of achieving consistent accuracy across continental and global-scale AGB maps, due to the absence of a comprehensive global reference dataset. Furthermore, \cite{Leonhardt2022} underscore the necessity for a dataset featuring globally distributed biomass reference records to improve regional studies. Similarly, \cite{Biomassters} advocate for the release of more openly accessible, deep learning-ready data. \cite{Tolan2024} highlight another aspect of this issue when they apply a correction based on GEDI to their recent 1-meter resolution canopy height maps for São Paulo (Brazil) and California (USA), aiming to compensate for the limited geographic diversity of their reference data. Taken together, these observations point at a critical gap in the literature: there is a need for a robust, globally diverse dataset to support high-resolution AGB mapping.

In their comprehensive review of deep learning for remote sensing of forest inventories, \cite{Hamedianfar2022} identify two critical avenues for progress. The first is the integration of data from multiple scales and sources, which has been shown to enhance predictive accuracy. For instance, incorporating diverse data modalities, such as DEM-derived Canopy Height Models (CHM) \cite{Puliti2020} and Synthetic Aperture Radar (SAR) data \cite{Liu2023}, not only alleviates issues of data saturation but also improves the predictive accuracy for high biomass values \cite{Zhang2019}. The second lever is the recognition and differentiation of various tree species, which can significantly refine the granularity and precision of forest inventory analyses. These strategies underscore the potential of deep learning to revolutionize forest monitoring and management through richer data integration and species-specific insights.

Our dataset fills in these gaps: it has a nominal resolution of $10\,$m, is representative of the global land cover and vegetation distribution, and assembles a range of data sources whose synergies have — to the best of our knowledge — not yet been systematically investigated for the task at hand.

\begin{table*}
\label{tab:existing-work}
\centering
\renewcommand{\arraystretch}{1.4} 
\begin{tabularx}{\textwidth}{@{}p{1cm}p{2.6cm}YYYYY@{}}
\toprule
 & \textbf{Method(s)} & \textbf{Geographical Extent} & \textbf{Resolution} & \textbf{Limitation} & \textbf{Reference data} & \textbf{Input data} \\
\midrule
\cite{Biomassters} & U-TAE, Swin U-Net TRansformer, U-Net++ & Finland & 10m & Localized & Airborne LiDAR and allometry & S1, S2 \\
\hline
\cite{Liu2023} & RF, SVM, CNN & Southwest China & 10m & Localized & Destructive AGB measurements & S1, S2 \\
\hline
\cite{Puliti2020} & RF & Northern Norway & 10m & Localized & Field measurements and allometry & ArcticDEM, S2 \\
\hline
\cite{SiyuLiu2023} & UNet & Europe & 30m & Localized & Field measurements, airborne LiDAR, allometry & PlanetScope, LiDAR \\
\hline
\cite{Leonhardt2022} & CGAN & Northwestern USA & 30m & Localized & US Carbon Monitoring System AGB map  & Landsat-8, ALOS-2	PALSAR-2 \\
\hline
\cite{Xu2017} & Maximum Entropy estimator & Democratic Republic of Congo & 100m & Localized, low-resolution & Airborne LiDAR and allometry & Landsat-8, ALOS-2	PALSAR-2, SRTM \\
\hline
\cite{Pascarella2023} & UNet & Vietnam, Myanmar & 100m & Localized, low-resolution & ESA CCI Map & S2 \\

\bottomrule
\end{tabularx}
\caption{Recent works and limitations.}
\label{tab:recent_work}
\end{table*}

\section{Dataset}
\label{sec:dataset}

In this section, we discuss the rationale for creating the dataset, detail its components, and describe the process of its generation.  For detailed information on data formats and downloading instructions, please refer to the supplementary material and the to-be-released project website.

\begin{figure}[ht]
    \centering
    \captionsetup{justification=centering,margin=0cm}
    \includegraphics[width=1.\linewidth]{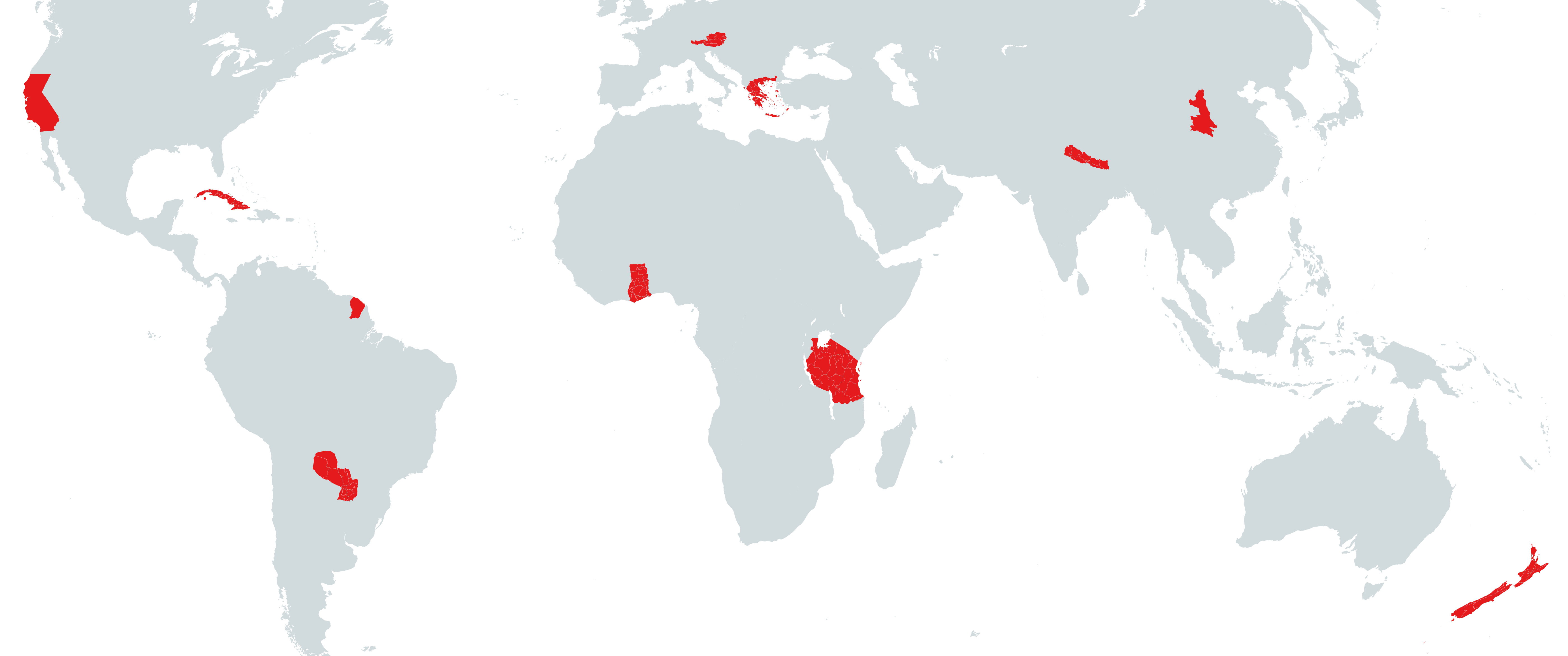}
    \caption{The regions of interest: California (USA), Cuba, Austria, Greece, Nepal, Shaanxi (China), French Guiana, Paraguay, Ghana, Tanzania, New Zealand.}
    \label{fig:subset}
\end{figure}

\subsection{Vegetation types analysis}
\label{sub:LC} 

\begin{figure}[th]
    \centering
    \captionsetup{justification=centering,margin=0cm}
    \includegraphics[width=1.\linewidth]{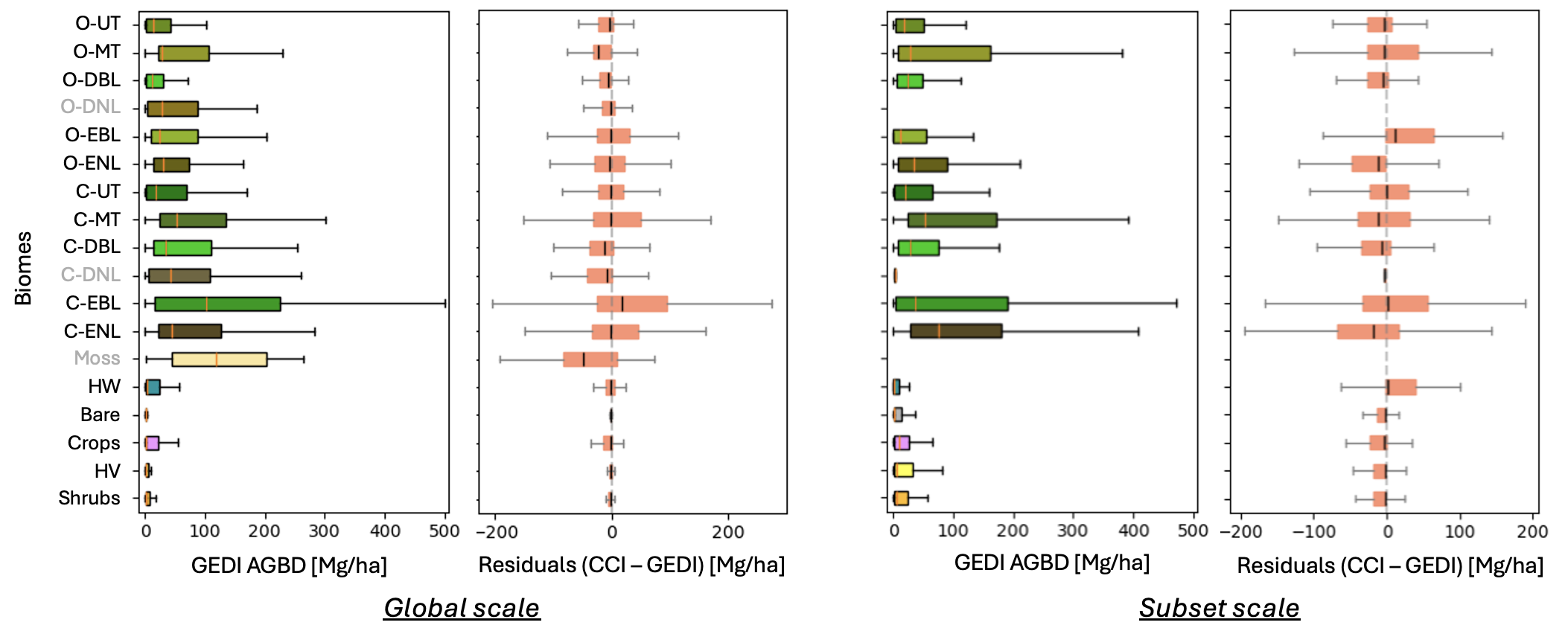}
    \caption{Per-biome distribution of GEDI AGBD values and ESA CCI residuals. \textit{Gray biomes are outside of GEDI coverage.}}
    \label{fig:cci-residuals}
\end{figure}

\paragraph{Land Cover Classification Map (LC).} To correctly represent the world's vegetation types, we turn to the Copernicus Global Land Service Dynamic Land Cover map \cite{LC2019}, which delivers a global map of the land cover at $100\,$m spatial resolution. It consists of a discrete classification system with 21 classes (following UN-FAO’s Land Cover Classification System) that differentiates between open (O) and closed (C) forest types, with the following sub-types: evergreen needle-leaved (ENL), deciduous needle-leaved (DNL), evergreen broad-leaved (EBL), deciduous broad-leaved (DBL), mixed type (MT), and unknown type (UT). Additional land cover classes include shrubland (Shrubs), herbaceous vegetation (HV), herbaceous wetland (HW), moss and lichen (Moss), bare or sparse vegetation (Bare), cropland (Crops), built-up areas (ignored), snow and ice (ignored), and permanent water bodies (ignored). We refer the reader to \cite{LC2019} for detailed descriptions of each class. We base our analysis on the $2019$ iteration of the product (latest available). Note that while the CGLS-LC100 map is global, we only consider vegetation types that are found within the GEDI coverage, as only there we have reference AGB data. As a consequence, vegetation types found only outside of the GEDI coverage are not accounted for (the grey biomes in Figure \ref{fig:cci-residuals}). These are: open and closed DNL only found in Siberia; and moss and lichen, mostly found in Northern Canada, Siberia and Greenland. 

\begin{figure}[th]
    \centering
    \captionsetup{justification=centering}
    \includegraphics[width=1\linewidth]{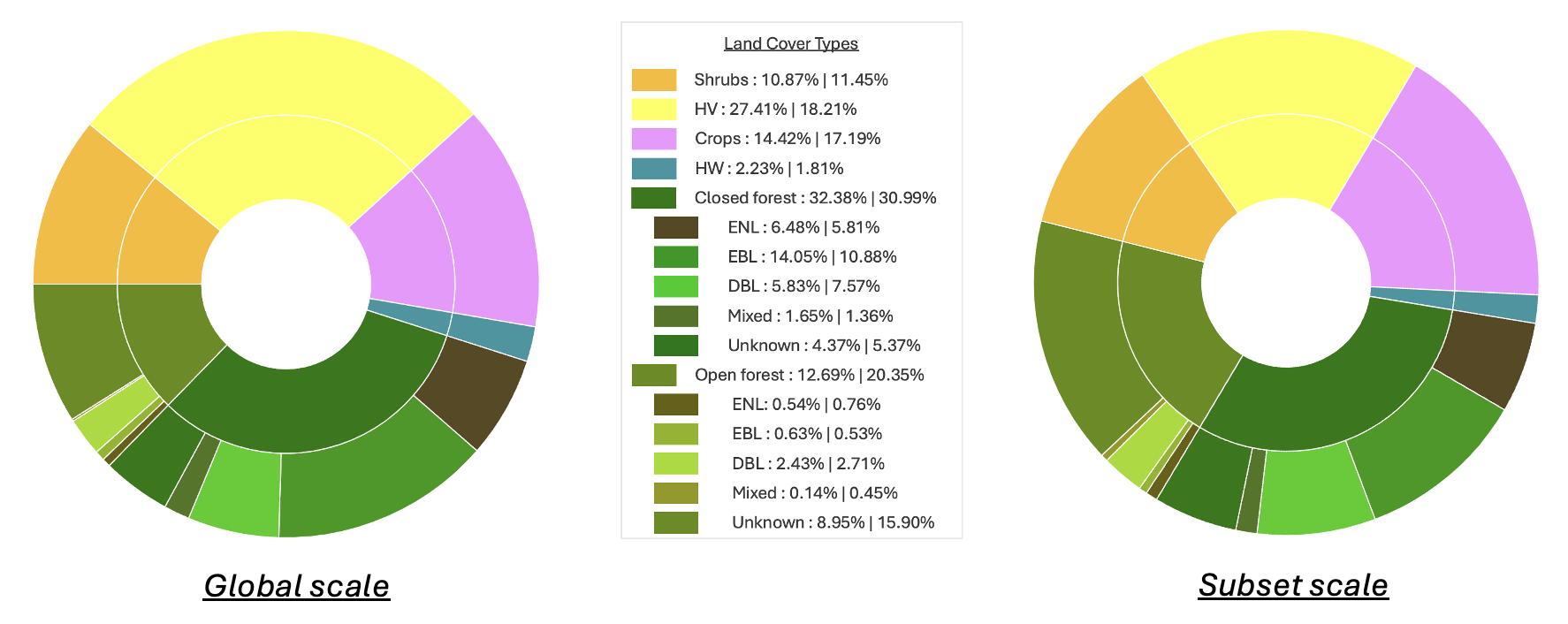}
    \caption{Land cover distribution across the world (left) and across our subset (right).}
    \label{fig:pie-charts}
\end{figure}

\paragraph{A globally representative vegetation distribution.} The \href{https://climate.esa.int/en/projects/biomass/}{ESA CCI map} is the most recent and adequate global AGB map available. It is provided by the European Space Agency (ESA), through its Climate Change Initiative (CCI) Biomass project. Global maps of AGB are provided for five epochs (2010, 2017, 2018, 2019 and 2020) at $100\,$m resolution. We observe that this map performs variably across different vegetation types, as shown in Figure \ref{fig:cci-residuals}. We utilise CCI AGB estimates and GEDI footprints from the year $2019$ to analyze residuals across various vegetation types within the GEDI coverage. Our analysis revealed significant variances, with underestimations in moss and lichen and overestimations in closed evergreen broadleaf forests, compelling us to adopt a design that accounts for differences between vegetation types. In the absence of a truly global dataset and facing the logistical barrier of large-scale remote sensing data downloads, we argue for a reasonably-sized dataset that mirrors the global vegetation distribution. To that end we select the following regions: California (USA), Cuba, Austria, Greece, Nepal, Shaanxi (China), French Guiana, Paraguay, Ghana, Tanzania and New Zealand. They are strategically chosen for their geographic and ecological diversity, as illustrated in Figures \ref{fig:subset} and \ref{fig:pie-charts}. This selection was based on a careful process. We chose moderate-sized countries or large administrative regions to avoid areas that were too large or too small. We considered the number of available GEDI footprints, as regions further from the equator have higher revisit times, affecting data density. To ensure diverse ecological representation, we used the Shannon entropy and Simpson's diversity index, based on land cover pixel values, to measure the biome diversity. Finally, regions were selected from various continents, with a balance above, below, and on the equator. Both the vegetation types and the ESA CCI residuals within our subset are in fairly good agreement with the global ones (Figure \ref{fig:cci-residuals}).

\subsection{Modalities}
\label{sub:modalities}

In the following we describe the data sources, how they were processed, and how we assembled them into the final dataset. An overview is provided in Figure \ref{fig:overview}. A more detailed explanation of each individual data field can be found in the Appendix (Table \ref{tab:dataset_overview} and Section \ref{sub:description}).

\textbf{GEDI}, an altimeter mission (2018-2023) on the International Space Station, produced high-resolution laser observations of the Earth's 3D structure. Each observation consists of a $25$m diameter footprint, with an associated AGB estimate. Due to its sampling pattern, GEDI has a (mostly) global yet sparse coverage: individual footprints from 51.6°N to 51.6°S. We downloaded all GEDI L4A\cite{GEDI_L4A} (v2.1) footprints for the regions of interest from the start of the mission until Dec.~31st 2020. We only kept biomass values in the range $[0,500]$ Mg/ha, as suggested by previous literature \cite{Carreiras2017}; we only retain the "power" beams, as previous work \cite{MTClem,Lahssini2022-fp} has found that they provide more reliable estimates than the "coverage" beams; and as suggested by the GEDI team, we discard footprints where \verb|sensitivity| $< 0.95$, \verb|l4_quality_flag| $\neq 1$, and \verb|selected_algorithm| $= 10$. Details on the process by which GEDI esimates the AGB can be found in the Appendix, in Section \ref{sec:gedi_agb}.

\textbf{Sentinel-2 (S2)}, a European multi-spectral imaging mission, delivers high-resolution imagery with a revisit frequency of $5$ days at the equator. We downloaded selected Level-1C products with less than 20$\%$ cloud coverage from October 2018 to December 2020 from \href{https://peps.cnes.fr/rocket/#/home}{PEPS}, and processed them to Level-2A using Sen2Cor \cite{Main-Knorn2017-ze} (v2.11). The pre-processing of the products involves the conversion of raw digital numbers ($DN$) to surface reflectance values via $SR = \frac{DN + \text{BOA}\_\text{offset} }{10,000}$, with $\text{BOA}\_\text{offset}$ equal to $0$ prior to January $2022$, and $-1000$ after that. We use all $12$ available spectral bands as input features, upsampling the $20$m and $60$m ones to $10$m resolution. We use the (upsampled) SCL band for patch selection purposes.

\textbf{PALSAR-2 (P2)}, on board the ALOS-2 satellite, is a Synthetic Aperture Radar (SAR) that can capture images at night and through clouds, making it a 24-hour, all-weather technology. The $25\,$m PALSAR-2 yearly mosaic\cite{Shimada_2014} is a seamless global image in WGS84 (EPSG:4326) projection, created by merging strips of SAR imagery for a year, featuring both HH and HV polarization backscatter. It uses a different gridding system from ESA, so for each year we download the necessary ALOS tiles, then mosaic, crop and reproject them to align with Sentinel-2 tiles. The provided digital numbers ($DN$) are linear backscatter amplitudes. Pre-processing involves converting DN values to $\gamma_0$ values in decibels via $ \gamma^0 = 10\cdot log_{10}(DN^2) - 83$.

\textbf{Digital Elevation Model (DEM)}. The ALOS Global Digital Surface Model (AW3D30)\cite{Tadono_2014} offers 30-meter resolution elevation data. As for P2 mosaics, we downloaded, mosaicked, and reprojected the DEM to align with Sentinel-2 tiles. 

\textbf{Land Cover (LC)}. We downloaded the Copernicus Global Land Service Dynamic Land Cover map (Section \ref{sub:LC}) at 100-meter resolution, upsampled, cropped and reprojected it to match the Sentinel-2 tiles. The map includes two layers: the discrete classification, and the probability of the discrete classification (which acts as a quality indicator).

\textbf{Canopy Height (CH)}. Lang et al. \cite{Lang2023} provide source code for canopy height mapping from Sentinel-2 imagery, using probabilistic deep learning. The model estimates CH for each input Sentinel-2 L2A image and merges redundant predictions from different dates to create a yearly CH map for each Sentinel-2 tile, along with a map of the associated standard deviations (STD). We follow this procedure to obtain yearly CH maps for 2019 and 2020 for all Sentinel-2 tiles within our regions of interest.

\textbf{Coordinates}. We additionally provide latitude and longitude information to the model. 

\begin{figure}[th]
    \centering
    \captionsetup{justification=centering,margin=0cm}
    \includegraphics[width=\linewidth]{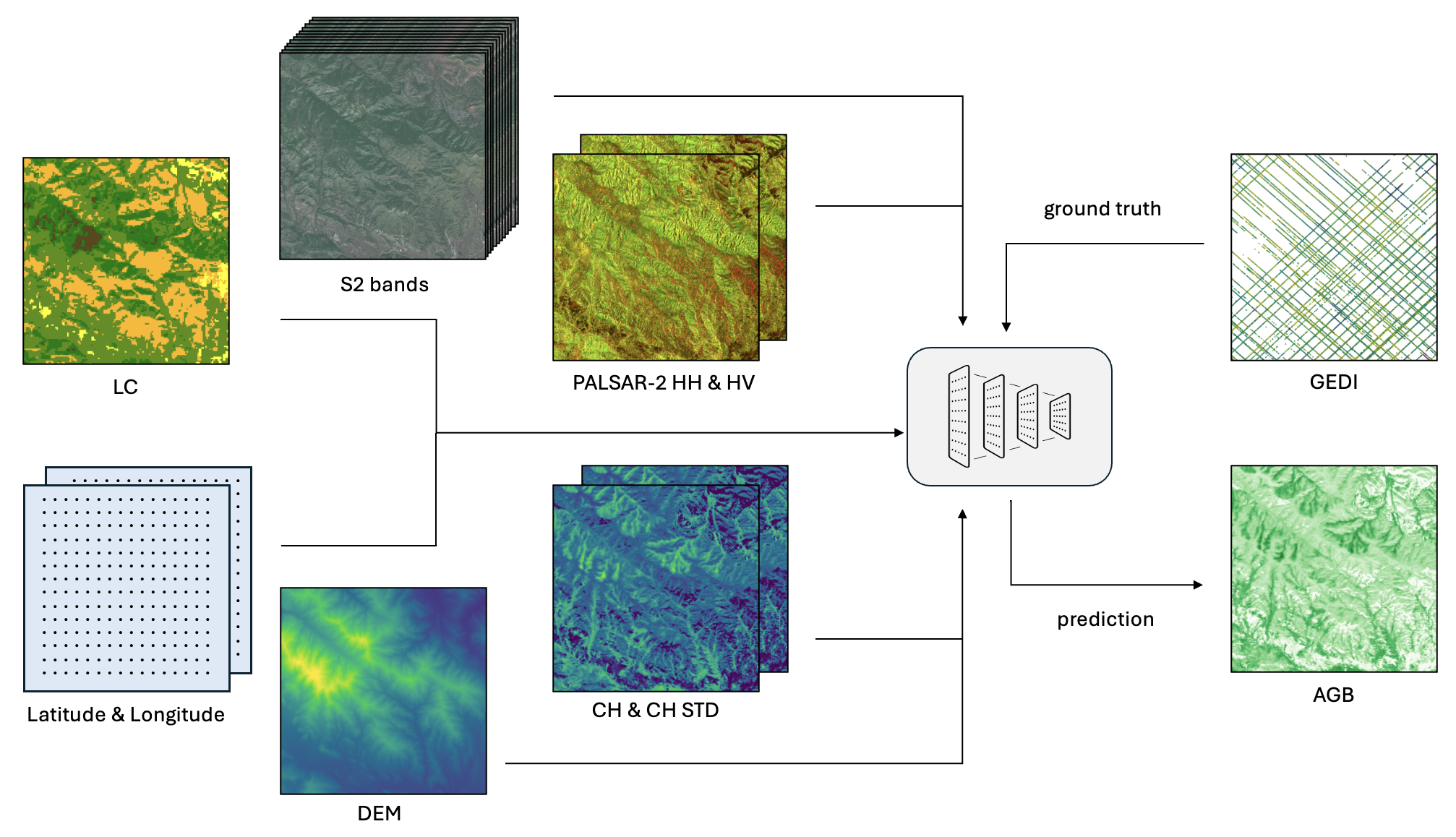}
    \caption{Visualization of all data sources along with our training pipeline. Note the sparse GEDI labels, which are upgraded to dense AGB maps by the prediction.}
    \label{fig:overview}
\end{figure}

\subsection{Dataset Assembly and Train/Test Split}
To create an analysis-ready dataset, we crop all data into \textit{patches} as follows. First, all data sources are up-sampled to a common resolution of $10\,$m, using nearest neighbor for categorical variables (the S2 SCL band, and the LC probability band) and bilinear interpolation for continuous variables (all others). We then iterate over each GEDI footprint, and fetch the corresponding data: for S2, the closest product within the six months preceding the footprint acquisition; for the yearly available data products (CH and P2), the data from the year of the footprint acquisition; and the product itself for non time-dependent data products (DEM and LC). Once all data sources are temporally aligned, we crop all raster data to 25$\times$25 pixel squares centered on the GEDI footprint, and store them in a HDF5 file.

The splitting into training, validation and test sets is not done at the level of GEDI footprints, but at the level of the Sentinel-2 tiles, meaning that all patches within the same $100\times100\,$km\textsuperscript{2} tile will belong to the same set. To align with our geographical distribution and vegetation types, we split each country/region into training, validation and test portions and aim to match the distribution of vegetation types within each portion to that of the whole region. To maintain our geographical distribution and vegetation types distribution, we perform a train/validation/test split for each region. We enforce that the vegetation types distribution within each region's split should match the whole region's distribution. This is done by randomly exploring the combinatorial assignment of tiles to the three splits until a sufficiently close match has been found. In each region $65\%$ of the tiles are used for training, $15\%$ for validation, and the remaining $20\%$ for testing. The detailed assignment per tile can be found on our project website.

\section{Experiments}
\label{sec:benchmark}

\subsection{Models}
\label{subsec:models}

Early works on biomass estimation employed linear regression models, but the relationship between satellite obervations and AGB is more complex and can be better modeled with high-capacity ML methods \cite{Li2020, Liu2023}. We provide a selection of pertinent ML models as benchmarks for the task.

\vspace{-0.4em}
\textbf{GBDT}. We start with a \verb|lightgbm| implementation of Gradient Boosted Decision Trees (GBDT). This model does not leverage the spatial context information of the patches but performs regression based only on the central pixel. The model was trained by minimising the RMSE loss, with early stopping enabled. Model hyper-parameters can be found in the Appendix. 

\vspace{-0.5em}
For the following deep learning methods, we always use a patch-wise training procedure: each patch (of size $25 \times 25$ or $15 \times 15$) has \textit{one} ground-truth pixel, its center. The model emits a prediction for each pixel in the patch, but only the central pixel prediction contributes to the loss. The optimisation of the network parameters is done in the standard way with mini-batch stochastic gradient descent, using the ADAM \cite{ADAM} variant and batch size 256.  Each model is trained five times with different random seeds to account for stochastic variability, and we report the mean RMSE and standard deviation over the five runs. All models run on consumer hardware, specifically we used NVIDIA GeForce RTX 2080 Ti GPUs available through our institution's high-performance cluster. The training pipeline is illustrated in Figure \ref{fig:overview}.

\vspace{-0.5em}
\textbf{FCN} ($0.5$M parameters). This model is a straightforward Fully Convolutional Neural Network (FCN) consisting of convolutional layers with batch normalization and ReLU activations. The channel depth increases as [32, 64, 128, 128, 128, 128]. Convolutional layers use a stride of 1 and padding of 1 to preserve spatial dimensions. The final convolution maps the 128 feature channels to a single output channel. 

\vspace{-0.5em}
\textbf{UNet} ($10$M parameters). A standard UNet implementation with an initial double convolution block, followed by symmetric downsampling and upsampling layers with skip connections to preserve spatial detail, and a final output convolution to produce a single-channel output. The number of down- and up-sampling layers depends on the patch size, with larger patches going through more layers.

\vspace{-0.5em}
\textbf{Lang et al.} ($13$M parameters). The architecture developed in \cite{Lang2023} for CH estimation, but trained (from scratch) for AGB estimation, with standard $\ell_2$-loss (rather than the original log-likelihood loss). It consists of a series of residual blocks with separable convolutions \emph{without} any downsampling. We use $728$ filters per block (instead of $256$).

\subsection{Results} 

All quantitative results are reported in Table \ref{tab:ablation}. The lowest error is achieved by the Lang et al. architecture, trained on all considered input channels but CH. The best models achieve RMSE values slightly below $60$ Mg/ha, consistent with previous literature. We elaborate on various aspects in the following paragraphs.

\textbf{Role of input features}. In order to study the impact of each data source, we conduct a series of experiments. We vary the input features available to each model: all features, only the canopy height (\textbf{CH}), only the $10\,$m RGB (B02, B03, B04) and NIR (B08) S2 bands (\textbf{RGBN}), all S2 bands (\textbf{S2 (all)}), all S2 bands plus the LC map (\textbf{LC}), all S2 bands plus the DEM (\textbf{DEM}), all S2 bands and the two ALOS-2 PALSAR-2 bands (\textbf{P2}), and all features except for the CH. Per-pixel latitude and longitude are always provided as inputs. In each setting, the $5$ models with the highest validation scores are applied to the held-out test set to obtain the mean RMSE values and associated standard deviations reported in Table \ref{tab:ablation}. The best model is the one which has access to additional input features (i.e., on top of S2); we attribute this to the comprehensive, complementary evidence that the model can rely on (namely optical, SAR, elevation, and vegetation).

\textbf{Patch size}. We consider patches of size consistent with the literature \cite{Lang2023, Tolan2024}. Namely, to study how varying the patch-size affects model performance, we train and evaluate with two different patch sizes: $15\times15$ and $25\times25$ pixels (see Table \ref{tab:ablation}). Our experiments showed minimal performance differences, favoring the $25\times25$ pixels patch size. Given the limited spatial context of most existing biomass retrieval methods, it is interesting that increasing the context window yields better results. At present it is unclear whether this is due to real context information or simply a data augmentation effect.

\begin{figure}[h]
    \centering
    \includegraphics[width=\linewidth]{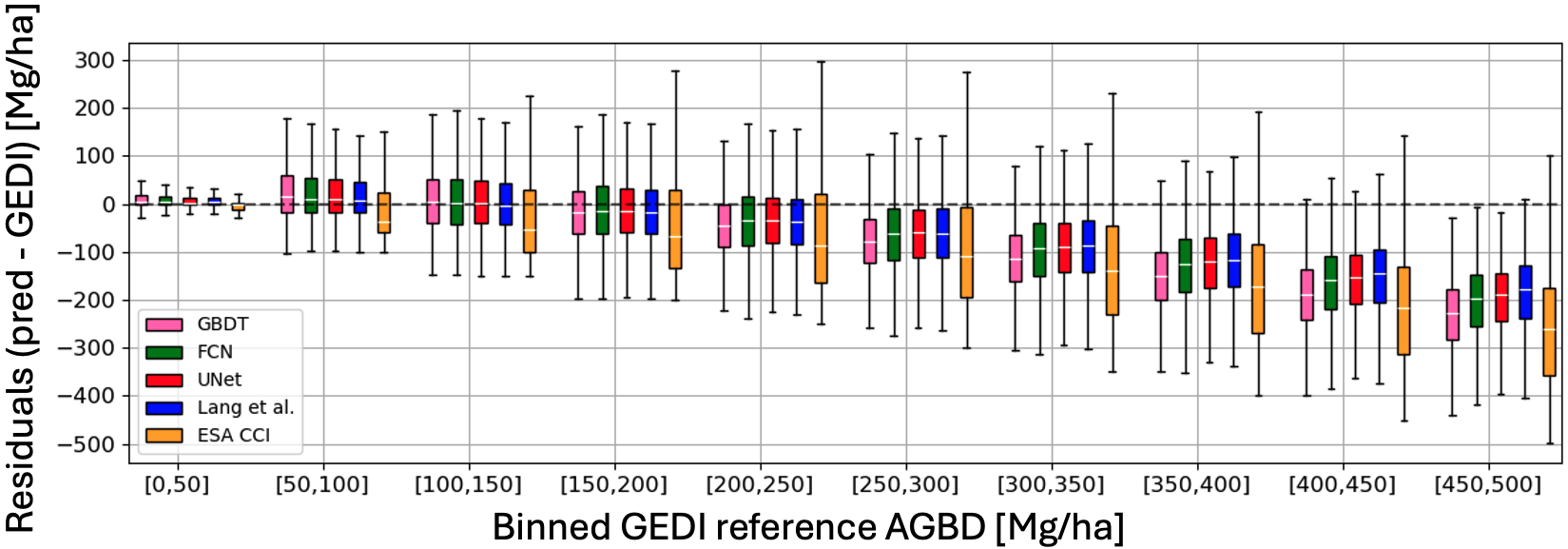}
    \captionsetup{justification=centering}
    \caption{Binned test residuals for the best-performing model of each architecture, and for the ESA CCI predictions.}
    \label{fig:residuals}
\end{figure}

\textbf{Binned residuals analysis}. Expanding on the previous analysis, we also analyse the residual distribution across various AGB intervals, see Figure \ref{fig:residuals}. Besides our baseline models described in Section \ref{subsec:models} we also include the ESA CCI AGB map, upsampled to $10\,$m resolution, in this analysis. That comparison is directly relevant, since ESA CCI also uses GEDI to calibrate their model, according to their Algorithm Theoretical Basis Document \cite[p.23]{ESACCI}. We point out that the distribution of AGB values is skewed: most ($50$\%) of all GEDI labels fall into the first bin ($[0,50]$).

\begin{figure}[h]
    \centering
    \includegraphics[width=\linewidth]{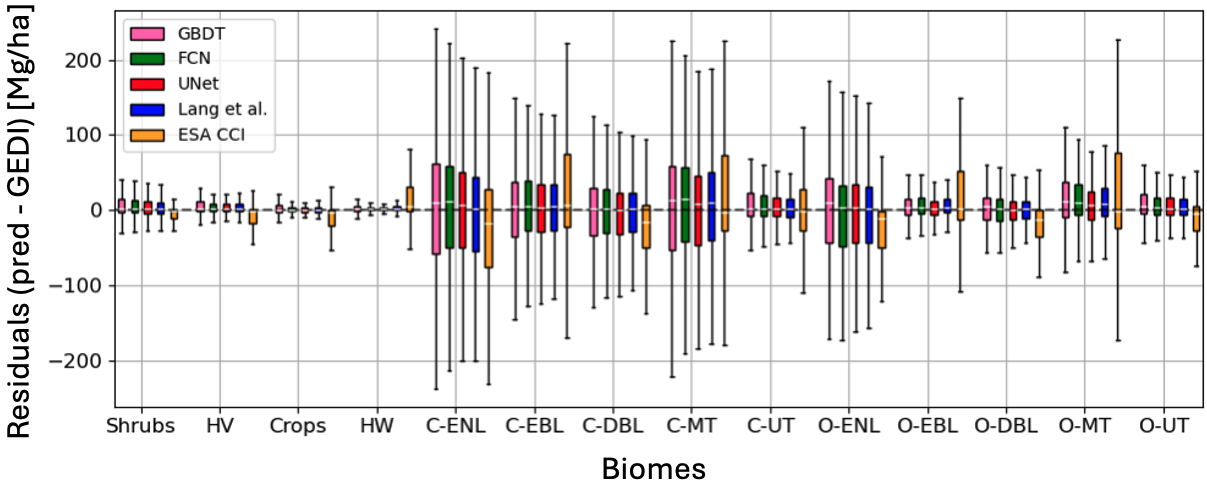}
    \captionsetup{justification=centering}
    \caption{Per-biome test residuals for the best-performing model of each architecture, and for the ESA CCI predictions.}
    \label{fig:biome_residuals}
\end{figure}

\textbf{Per biome residuals analysis}. To emphasize how much AGB estimates depend on the biome at hand, we plot the residuals across the biomes, in Figure \ref{fig:biome_residuals}.

\begin{figure}[h]
    \centering
    \includegraphics[width=1.0\linewidth]{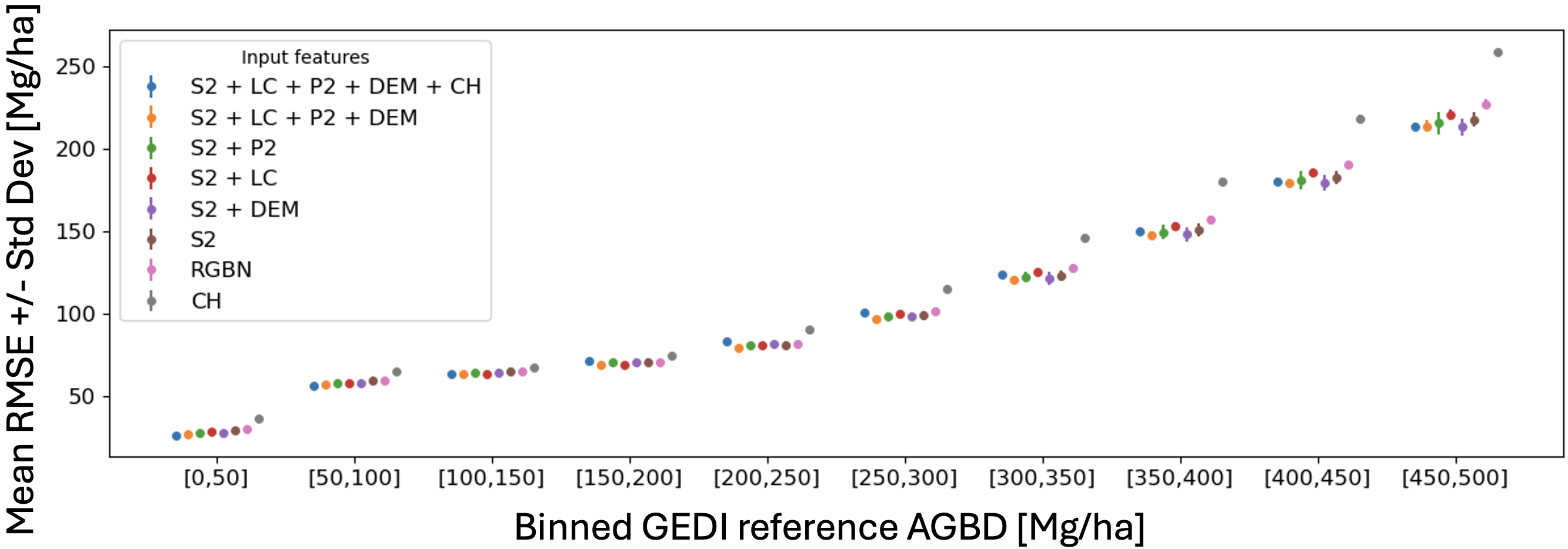}
    \captionsetup{justification=centering}
    \caption{Binned mean test RMSE ($\downarrow$) and associated standard deviation per model, with various inputs.}
    \label{fig:saturation}
\end{figure}

\textbf{Saturation analysis}. A known issue for AGB estimation is \textit{saturation}, which happens when beyond a certain AGB threshold, the RS information no longer reflects the change in AGB, making it challenging to estimate high values. To investigate whether including additional data sources mitigates with this issue, we include the binned residuals of the Lang et al. models trained on the various feature combinations. The results are shown in Figure \ref{fig:saturation}, and indicate that having access to the additional features makes the AGB estimates better particularly in the higher bins.

\begin{table*}[th]
\centering
\renewcommand{\arraystretch}{1.2} 
\resizebox{\textwidth}{!}{
\begin{tabular}{ccccccccccccccc}
\toprule
\multicolumn{6}{c}{\textbf{Features}} & \multicolumn{7}{c}{\textbf{Methods}}\\
\cmidrule(lr){1-6} 
\cmidrule(lr){7-13} 
\textbf{CH} & \textbf{RGBN} & \textbf{S2 (all)} & \textbf{LC} & \textbf{DEM} & \textbf{P2} & \textbf{GBDT} & \multicolumn{2}{c}{\textbf{FCN}} & \multicolumn{2}{c}{\textbf{UNet}} & \multicolumn{2}{c}{\textbf{Lang et al.}} \\
 & & & & & & \textbf{(1×1)}& \textbf{(15×15)} & \textbf{(25×25)} & \textbf{(15×15)} & \textbf{(25×25)} & \textbf{(15×15)} & \textbf{(25×25)} \\
\midrule

× & & & & & & \cellcolor[RGB]{254,186,186}\shortstack{70.73 \\ $\pm$0.18} & \cellcolor[RGB]{254,196,196}\shortstack{69.29 \\ $\pm$0.16} & \cellcolor[RGB]{254,196,196}\shortstack{69.33 \\ $\pm$0.07} & \cellcolor[RGB]{254,197,197}\shortstack{69.08 \\ $\pm$0.08} & \cellcolor[RGB]{254,200,200}\shortstack{68.54 \\ $\pm$0.12} & \cellcolor[RGB]{254,200,200}\shortstack{68.54 \\ $\pm$0.27} & \cellcolor[RGB]{254,202,202}\shortstack{68.11 \\ $\pm$0.24} \\ \midrule
& × & & & & & \cellcolor[RGB]{255,155,155}\shortstack{72.92 \\ $\pm$0.18} & \cellcolor[RGB]{254,214,214}\shortstack{64.97 \\ $\pm$0.25} & \cellcolor[RGB]{254,214,214}\shortstack{64.75 \\ $\pm$0.11} & \cellcolor[RGB]{254,219,219}\shortstack{63.14 \\ $\pm$0.29} & \cellcolor[RGB]{229,229,255}\shortstack{61.97 \\ $\pm$0.09} & \cellcolor[RGB]{227,227,255}\shortstack{61.60 \\ $\pm$0.13} & \cellcolor[RGB]{216,216,255}\shortstack{60.39 \\ $\pm$0.19} \\ \midrule
& & × & & & & \cellcolor[RGB]{254,195,195}\shortstack{69.41 \\ $\pm$0.11} & \cellcolor[RGB]{254,218,218}\shortstack{63.52 \\ $\pm$0.22} & \cellcolor[RGB]{254,218,218}\shortstack{63.48 \\ $\pm$0.08} & \cellcolor[RGB]{230,230,255}\shortstack{62.00 \\ $\pm$0.09} & \cellcolor[RGB]{219,219,255}\shortstack{60.69 \\ $\pm$0.05} & \cellcolor[RGB]{216,216,255}\shortstack{60.36 \\ $\pm$0.13} & \cellcolor[RGB]{201,201,255}\shortstack{59.15 \\ $\pm$0.22} \\ \midrule
& & × & × & & & \cellcolor[RGB]{254,209,209}\shortstack{66.48 \\ $\pm$0.05} & \cellcolor[RGB]{254,221,221}\shortstack{62.40 \\ $\pm$0.19} & \cellcolor[RGB]{254,220,220}\shortstack{62.47 \\ $\pm$0.12} & \cellcolor[RGB]{222,222,255}\shortstack{61.09 \\ $\pm$0.18} & \cellcolor[RGB]{212,212,255}\shortstack{59.99 \\ $\pm$0.12} & \cellcolor[RGB]{208,208,255}\shortstack{59.65 \\ $\pm$0.10} & \cellcolor[RGB]{196,196,255}\shortstack{58.82 \\ $\pm$0.14} \\ \midrule
& & × & & × && \cellcolor[RGB]{254,204,204}\shortstack{67.57 \\ $\pm$0.06} & \cellcolor[RGB]{254,221,221}\shortstack{62.43 \\ $\pm$0.24} & \cellcolor[RGB]{228,228,255}\shortstack{61.78 \\ $\pm$0.69} & \cellcolor[RGB]{217,217,255}\shortstack{60.47 \\ $\pm$0.67} & \cellcolor[RGB]{204,204,255}\shortstack{59.33 \\ $\pm$0.40} & \cellcolor[RGB]{192,192,255}\shortstack{58.55 \\ $\pm$0.19} & \cellcolor[RGB]{178,178,255}\shortstack{57.90 \\ $\pm$0.21} \\ \midrule
& & × & & & × & \cellcolor[RGB]{254,206,206}\shortstack{67.27 \\ $\pm$0.08} & \cellcolor[RGB]{229,229,255}\shortstack{61.95 \\ $\pm$0.19} & \cellcolor[RGB]{228,228,255}\shortstack{61.81 \\ $\pm$0.18} & \cellcolor[RGB]{218,218,255}\shortstack{60.59 \\ $\pm$0.14} & \cellcolor[RGB]{203,203,255}\shortstack{59.27 \\ $\pm$0.11} & \cellcolor[RGB]{201,201,255}\shortstack{59.12 \\ $\pm$0.11} & \cellcolor[RGB]{186,186,255}\shortstack{58.22 \\ $\pm$0.23} \\ \midrule
& & × & × & × & ×& \cellcolor[RGB]{254,213,213}\shortstack{65.16 \\ $\pm$0.02} & \cellcolor[RGB]{221,221,255}\shortstack{60.97 \\ $\pm$0.05} & \cellcolor[RGB]{221,221,255}\shortstack{60.97 \\ $\pm$0.19} & \cellcolor[RGB]{210,210,255}\shortstack{59.84 \\ $\pm$0.10} & \cellcolor[RGB]{191,191,255}\shortstack{58.50 \\ $\pm$0.16} & \cellcolor[RGB]{184,184,255}\shortstack{58.16 \\ $\pm$0.15} & \cellcolor[RGB]{155,155,255}\shortstack{57.14 \\ $\pm$0.29} \\ \midrule
× &  & × & × & × & × & \cellcolor[RGB]{254,217,217}\shortstack{63.76 \\ $\pm$0.09} & \cellcolor[RGB]{221,221,255}\shortstack{60.98 \\ $\pm$0.23} & \cellcolor[RGB]{221,221,255}\shortstack{60.92 \\ $\pm$0.23} & \cellcolor[RGB]{214,214,255}\shortstack{60.18 \\ $\pm$0.36} & \cellcolor[RGB]{188,188,255}\shortstack{58.33 \\ $\pm$0.47} & \cellcolor[RGB]{182,182,255}\shortstack{58.04 \\ $\pm$0.31} & \cellcolor[RGB]{168,168,255}\shortstack{57.52 \\ $\pm$0.41} \\ 

\bottomrule
\end{tabular}
}
\captionsetup{justification=centering,margin=0cm}
\caption{Mean test RMSE ($\downarrow$) and associated standard deviation per model, with various inputs. Crosses denote the \emph{presence} of a feature, values in brackets denote patch size. Values are colored from dark red (higher) to dark blue (lower).
\label{tab:ablation}}
\end{table*}

\begin{figure*}[h]
    \centering
\captionsetup{justification=centering,margin=0cm}
    \includegraphics[width=\linewidth]{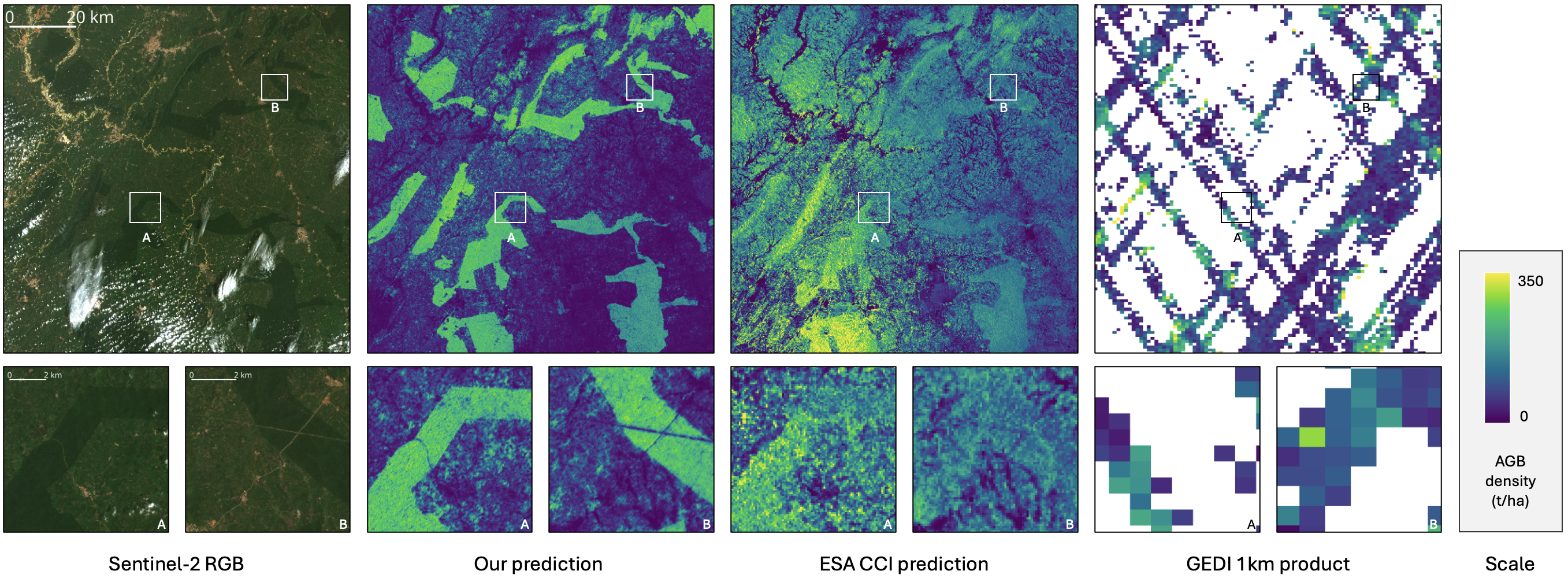}
    \caption{Sentinel-2 tile 30NXM (Ghana), our best model's prediction, the corresponding ESA CCI map, and GEDI L4B product. Global view (top) and zoomed in views (bottom).}
    \label{fig:visualization}
\end{figure*}

\textbf{Qualitative results inspection}. For a visual impression, Figure \ref{fig:visualization} shows the predictions of the best-performing model versus the CCI map for a selected Sentinel-2 tile (30NXM, located in Ghana). As the GEDI L4A reference footprints are too small and sparse to visualize, we include the GEDI L4B 1km product \cite{GEDI_L4B}, which provides statistically inferred estimates of mean AGB from the footprints located within the borders of each 1 km cell. This figure illustrates how the high spatial resolution of our estimates is able to preserve spatial details that are not captured in the CCI product, meaning that they would not be restored by conventional upscaling of existing maps.

\section{Discussion}
\label{sec:discussion}

\textbf{Results}. Our various analyses indicate that including input features beyond Sentinel-2 is beneficial for the task of AGB estimation, and that more complex models tend to better capture the complex relationship between input features and AGB. 

\textbf{Stand-alone task}. As AGB is highly dependent on CH (via allometric equations), one could wonder whether CH information is enough to estimate AGB. We conclude that it is not the case: our results show that learning AGB "from scratch" (using S2, P2, LC and DEM) yields better results than learning from CH directly. While this applies to the specific Lang et al. CH product at hand, the most recent ones published \cite{Pauls2024, Tolan2024} also rely on GEDI CH estimates, so we expect similar results. We thus argue that AGB estimation should be tackled as its own task.

\textbf{Limitations}. Despite the nominal 10$\,$m resolution of our estimates, the effective spatial resolution at which biomass is identified is lower. Additionally, the uncertainty in the footprint-level GEDI AGBD product used as reference data influences the accuracy and uncertainties of the model predictions. We elaborate on these points in the Appendix, Section \ref{sec:gedi_limitations}.

\section{Conclusion}
We provide a ML-ready, unrestricted, easily accessible dataset for high-resolution biomass estimation from remote sensing data. This dataset consists of $\approx 16\cdot10^6$ patches, covering a diverse and representative selection of geographic locations with a total of $\approx500\cdot10^6$ hectares and two different years ($2019$ and $2020$). Furthermore, we complement the dataset with a suite of baseline architectures and trained weights for others to build upon. We hope that our dataset may help to reduce the geographical and ecosystem-related biases of biomass models, enhance the accuracy of future biomass maps, and perhaps contribute towards a regular, operational high-resolution monitoring system.

\section{Acknowledgments}
\label{sec:acknowledgments}
The data products used for this paper have been collected from various sources:
\vspace{-0.5em}
\begin{myitemize}
    \item the Global PALSAR-2/PALSAR/JERS-1 Mosaics and Forest/Non-Forest Maps of the Japan Aerospace Exploration Agency;
    \item the ALOS Global Digital Surface Model "ALOS World 3D - 30m (AW3D30)" of the Japan Aerospace Exploration Agency;
    \item the Land Cover 2019 (v3) of the European Union's Copernicus Land Monitoring Service;
    \item the Sentinel-2 L1C products, courtesy of Copernicus, downloaded via the French \href{https://peps.cnes.fr/rocket/#/home}{PEPS} platform;
    \item the GEDI L4A product, courtesy of the NASA GEDI Team, downloaded from Oak Ridge National Laboratory DAAC.
\end{myitemize}
We thank them for making this data openly available for academic purposes.

\printbibliography


\clearpage

\section{Appendix}

\subsection{Dataset description}
\label{sub:description}

Each sample in the dataset contains a pair of pre-cropped images (\textit{patches}) and their corresponding AGB labels. You can find an up-to-date description of the dataset on \href{https://huggingface.co/datasets/prs-eth/AGBD}{HuggingFace}.

We hereby provide more information on the encoding of certain features:
\begin{itemize}

    \item Geographical Coordinates — we perform sine/cosine encoding, as follows: \\
    \verb|lat_cos| $= (cos(\pi * lat / 90) + 1)/2$ \\
    \verb|lat_sin| $= (sin(\pi * lat / 90) + 1)/2$ \\
    \verb|lon_cos| $= (cos(\pi * lon / 180) + 1)/2$ \\
    \verb|lon_sin| $= (sin(\pi * lon / 180) + 1)/2$
    
    \item Land Cover Information — we perform sine/cosine encoding, as follows: \\
    \verb|lc_cos| $=(cos(2\pi * lc\_class / 100) + 1) / 2$ \\
    \verb|lc_sin| $=(sin(2\pi * lc\_class / 100) + 1) / 2$ \\
    where \verb|lc_class| is the discrete classification of the land cover. 

\end{itemize}

Additionally from the features described in this manuscript, users of the dataset have access to the following features (mainly GEDI L4A metadata):

\begin{itemize}
    \item AGBD Standard Error: The uncertainty estimate associated with the aboveground biomass density prediction for each GEDI footprint.
    
    \item Elevation: The height above sea level at the location of the GEDI footprint.
    
    \item Leaf-Off Flag: Indicates whether the measurement was taken during the leaf-off season, which can impact canopy structure data.
    
    \item Plant Functional Type (PFT) Class: Categorization of the vegetation type by GEDI (e.g., deciduous broadleaf, evergreen needleleaf).
    
    \item Region Class: The geographical area where the footprint is located (e.g., North America, South Asia).
    
    \item RH98 (Relative Height at 98\%): The height at which 98\% of the returned laser energy is reflected, a key measure of canopy height.
    
    \item Sensitivity: The proportion of laser pulse energy reflected back to the sensor, providing insight into vegetation density and structure.
    
    \item Solar Elevation: The angle of the sun above the horizon at the time of measurement, which can affect data quality.
    
    \item Urban Proportion: The percentage of the footprint area that is urbanized, helping to filter or adjust biomass estimates in mixed landscapes.
    
    \item Date of GEDI Footprints: The specific date on which each GEDI footprint was captured, adding temporal context to the measurements.
    
    \item Date of Sentinel-2 Image: The specific date on which each Sentinel-2 image was captured, ensuring temporal alignment with GEDI data.
\end{itemize}

\subsection{Hosting, licensing, and maintenance plan}

The processed, ready-for-use data is hosted on \href{https://huggingface.co/datasets/prs-eth/AGBD}{HuggingFace}, where it can be easily accessed:
\begin{figure}[ht]
\centering
\includegraphics[width=\linewidth]{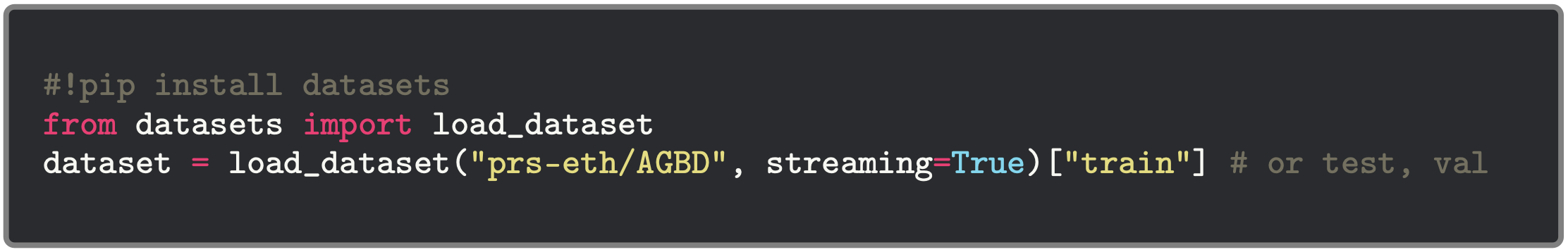}
\end{figure}

The raw data, dense predictions, and model weights are hosted on the \href{https://www.research-collection.ethz.ch/handle/20.500.11850/674193}{ETH Research Collection}, guaranteeing long-term preservation. 

The data is available under a \href{https://creativecommons.org/licenses/by-nc/4.0/}{CC BY-NC 4.0 DEED} license. You are free to copy and redistribute the material in any medium or format; remix, transform, and build upon the material. You must give appropriate credit, provide a link to the license, and indicate if changes were made. Importantly, you may not use the material for commercial purposes.

The dataset will be maintained by the authors to the best of their abilities and availability. 

\begin{table*}[h]
\centering
\renewcommand{\arraystretch}{1.5} 
\begin{tabularx}{\textwidth}{@{}p{2cm}YYYYYY@{}}
\toprule
 & \textbf{GEDI} & \textbf{Sentinel-2 (S2)} & \textbf{ALOS-2 PALSAR-2 (P2)} & \textbf{ALOS Digital Elevation Model (DEM)} & \textbf{Land Classification (LC)} & \textbf{CH} \\
\midrule
\textbf{Source} & NASA & Copernicus & JAXA & JAXA & Copernicus & Lang et al. \\
\hline
\textbf{Resolution (m)}  & 25 & [10,20,60] & 25 & 30 & 100 & 10 \\
\hline
\textbf{\# of channels} & - & 12 & 2 & 1 & 2 & 2 \\
\hline
\textbf{Temporality} & - & $\approx$ every 5 days & yearly & once & yearly & yearly \\
\bottomrule
\end{tabularx}
\caption{Data sources \label{tab:dataset_overview}}
\end{table*}

\begin{figure}[H]
    \centering
    \captionsetup{justification=centering}
    \includegraphics[width=\linewidth]{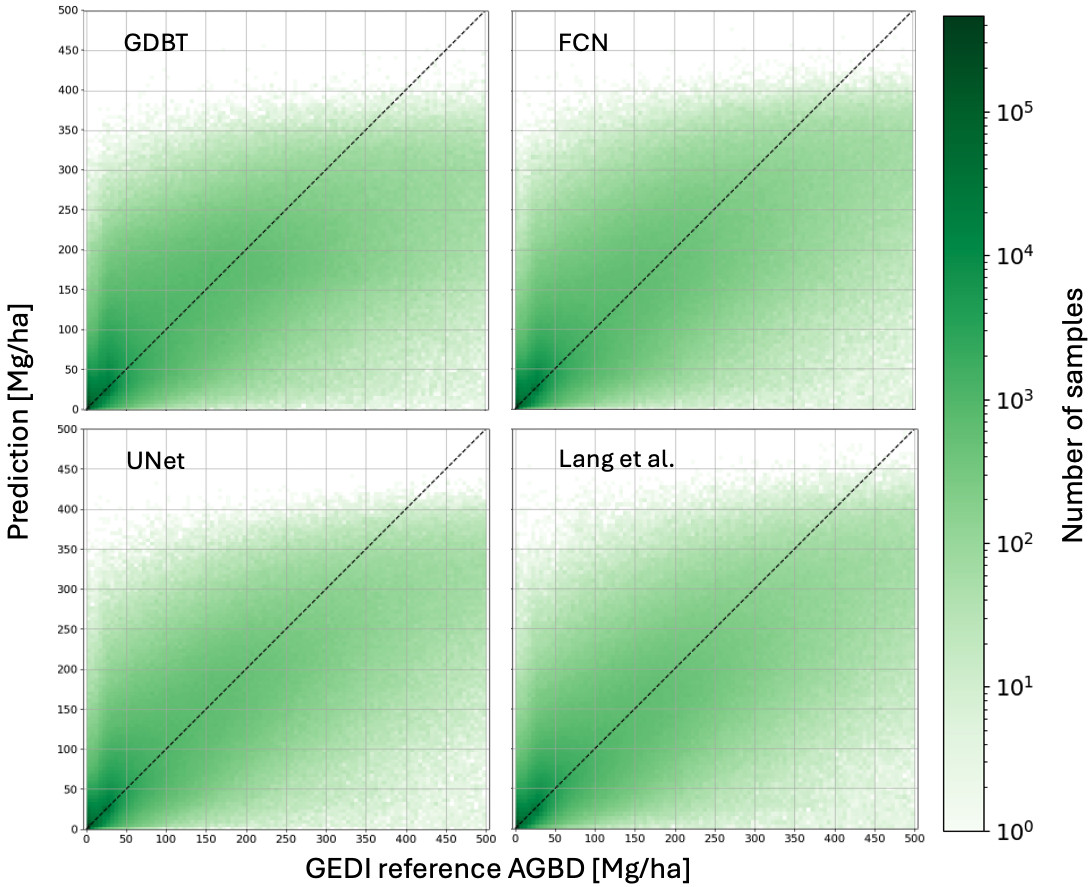}
    \caption{Density performance plots for the best model of each architecture.}
    \label{fig:density}
\end{figure}

\textbf{Confusion density plots}. We provide the confusion density plots for the best performing model of each architecture, in Figure \ref{fig:density}. It shows good agreement between the predictions and GEDI reference, and provides insights on the distribution of the biomass values.

\subsection{GBDT model initialization parameters}
\label{sec:params}

We hereby specify the initialization parameters for the GBDT models: \\
\verb|num_leaves=165, max_depth=-1, learning_rate=0.1|, \verb|num_iterations = 1000, min_split_gain = 0.|, \verb|min_child_weight =1e-3, min_child_samples=20|, \verb|reg_alpha= 0, reg_lambda = 5|, \verb|bagging_fraction = 0.5, bagging_freq = 1|.

\subsection{Dense predictions}
We generate the dense $10$m resolution AGB estimates for all Sentinel-2 tiles in the dataset coverage. For each tile, we pick the Sentinel-2 product that has the least cloud coverage over all products available, and perform inference on that product. As such, our dense predictions span multiple years. On the \href{https://www.research-collection.ethz.ch/handle/20.500.11850/674193}{ETH Research Collection}, the file \verb|Inference > mapping.pkl| indicates which specific product was used for inference for each tile. Note that this was a design choice, and one can generate dense predictions for specific years.

\subsection{GEDI AGB Estimates}
\label{sec:gedi_agb}

The GEDI instrument uses a LiDAR laser to repeatedly scan a 25m surface footprint, where each measurement consists of a waveform from the laser's returned signal. From the cumulative return energy, the GEDI team identifies the ground return, which is used as the reference for the relative height (RH) metrics. The RH[X] metrics correspond to the relative height at which [X] percent of the total accumulated energy is returned. From the RH metrics, the GEDI team estimates biomass. To do so, they developed their own allometric equations, based on the Plant Functional Type (PFT), the region of the world, and the RH metrics. In order to calibrate them, they leveraged world-wide crowd-sourced data. Coincident airborne LiDAR and ground plot field inventory were put together. The airborne LiDAR data was used to simulate GEDI-like waveforms and derived RH metrics. These simulated RH metrics were then used to calibrate the biomass equations with the field biomass values as reference data. Further information can be found in the Algorithm Theoretical Basis Document\cite{Kellner2023}.

\subsection{GEDI Limitations}
\label{sec:gedi_limitations}

\textbf{Resolution}. The GEDI mission operates with a $25$-meter diameter footprint on the ground. The laser pulse from GEDI's LiDAR system follows a Gaussian intensity profile, with the highest precision at the center of the footprint, gradually decreasing toward the edges \cite{Hancock2019}. Within this $25$-meter footprint, GEDI measures the highest reflecting surface, which is then used to derive Relative Height metrics, from which they estimate the AGB. We then chose to rasterize these footprints at a 10-meter resolution, which is a well-established method as supported by the literature \cite{Pauls2024, Lang2023}, and aligns with the highest resolution available among the sensors we considered, such as the $10$-meter bands from Sentinel-2. 

\textbf{Uncertainty}. The uncertainty in the footprint-level GEDI AGBD product used as reference data influences the accuracy and uncertainties of the model predictions. Future work could incorporate field measurements or airborne LiDAR data. However, given the sparsity of GEDI measurements, co-registering these with in-situ sparse measurements is challenging. Notably, in order to assess the quality of GEDI AGB estimates, the authors of \cite{Hunka2023} had to rely on an aggregated 1km x 1km grid GEDI product (where the samples present within the borders of each 1 km cell are used to statistically infer mean AGBD). Our denser and higher resolution estimates provide a crucial step forward, as well-calibrated estimates can be used as a proxy to assess the uncertainties in GEDI data by comparing them in-situ measurements. We believe that this will ultimately need a long-term community initiative across ecology, forestry, remote sensing and ML.

\textbf{Geolocation error}. While the geolocation error was an issue with the first version of the data released by GEDI, they released an updated version of their dataset (that we are using) with corrected geolocation errors. Notably, the authors of \cite{Tang2023} investigated the geolocation error of the updated version, and “did not find any significant relationship between geolocation error and performance”. They find that 80.8\% of the footprints have a geolocation error strictly smaller than 10m.

\textbf{Sampling pattern} Another point to keep in mind is that the spatial distribution of GEDI footprints is not uniform: the sampling gets progressively denser as one approaches towards the borders of the observed region at $\pm51.6^\circ$. 

\end{document}